\documentclass[11pt,a4paper]{article}

\usepackage[utf8]{inputenc}
\usepackage[T1]{fontenc}
\usepackage{lmodern}
\usepackage[margin=2.5cm]{geometry}
\usepackage{amsmath}
\usepackage{graphicx}
\usepackage{booktabs}
\usepackage{multirow}
\usepackage{caption}
\usepackage{enumitem}
\usepackage{authblk}
\usepackage[numbers,sort&compress]{natbib}
\usepackage[hidelinks]{hyperref}
\usepackage{setspace}
\graphicspath{{figures/}}
\newcommand{\Htwo}{H\textsubscript{2}}

\title{Leakage-Safe Machine Learning for Hydrogen Embrittlement Detection in 316L Stainless Steel: A Region-Held-Out Evaluation of Texture and Deep Features in SEM Micrographs}

\author[a,b,*]{Muhammad Awais}
\author[b]{Muhammad Yaseen}
\author[b]{Abdul Shakoor}
\author[c]{Niaz Ahmed Niaz}
\author[b]{Huria Zia}
\author[b]{Muhammad Zain Shakoor}

\affil[a]{Dipartimento di Fisica e Astronomia ``G. Galilei'', Universit\`a di Padova, Italy}
\affil[b]{Institute of Physics, Bahauddin Zakariya University, Multan, Pakistan}
\affil[c]{Institute of Physics, Bahauddin Zakariya University, Multan, Pakistan}
\affil[*]{Corresponding author: \href{mailto:muhammad.awais@phd.unipd.it}{muhammad.awais@phd.unipd.it}}

\date{}

\begin{document}

\maketitle

\begin{abstract}
Scanning electron microscopy (SEM) is routinely used to characterize the microstructural changes caused by hydrogen embrittlement (HE) in structural steels. Machine learning can automate this characterization, but models are often evaluated on image-level splits. When several images come from the same specimen region, such splits leak information between training and test sets. Here we propose a region-held-out protocol for classifying as-received (AR) and hydrogen-charged (\Htwo) SEM micrographs of 316L stainless steel, based on Leave-One-Region-Out (LORO) cross-validation over 14 spatial regions (8 AR, 6 \Htwo; 31 images). We compared six feature-classifier combinations built on local binary patterns (LBP), grey-level co-occurrence matrices (GLCM), self-supervised convolutional embeddings pretrained on 143 unlabeled SEM images, and a convolutional neural network (CNN). The simplest texture approach, LBP with a support vector machine (LBP+SVM), performed best: balanced accuracy 0.79, \Htwo{} recall 0.69 and \Htwo{} precision 0.82, ahead of every deep-learning and combined-feature model. A group-level permutation test (500 permutations sampled from the 3003 possible region-to-label assignments) gave $p = 0.008$, so the result cannot be explained by a chance alignment of the region structure. Grad-CAM maps from a CNN trained on the full dataset tended to concentrate on localized surface and grain-boundary features, where hydrogen-induced morphological changes are known to occur. Under a leakage-safe, statistically validated protocol, texture descriptors recover a hydrogen-charging signature from SEM micrographs even with few samples, and the same protocol can be carried over to larger HE detection studies in other alloy systems.
\end{abstract}


\section{Introduction}

A green hydrogen (\Htwo) economy needs infrastructure that stays stable and reliable for decades, from production through storage and transport. Stainless steels are the leading candidate materials, and AISI 316L in particular combines high strength, good ductility and fatigue resistance. These alloys are nonetheless vulnerable to one degradation mechanism above all. In hydrogen embrittlement (HE), atomic hydrogen enters the metal matrix, interacts heterogeneously with the microstructure and lowers the fracture toughness considerably. The interaction can alter localized deformation and ease dislocation motion, and the failure mode shifts from safe ductile fracture to catastrophic brittle fracture. Hydrogen-defect interactions below the surface are complex and hard to measure directly. In practice, engineering failure assessments therefore infer the onset of HE from the topographical and microstructural signatures it leaves in the material~\cite{robertson2015,yu2024}.

High-resolution scanning electron microscopy (SEM) is the main tool for capturing these visual signs of hydrogen damage. Manual microstructural analysis, however, is slow and open to human bias. Automated computer vision and machine learning models scale much better, but they are only useful in practice if they generalize across spatially varying microstructural domains instead of memorizing the grain textures of particular regions. This paper addresses that gap with a region-held-out evaluation of SEM-based hydrogen embrittlement detection in 316L. Distinct microstructural regions are systematically isolated during model training, so each model is tested on a zone it has never seen. The aim is automated diagnostics that pick up general, robust physical signatures, which is what the safe use of 316L components in green hydrogen technologies requires~\cite{hatano2014}.

Materials science and engineering are becoming increasingly data-driven, with high-throughput imaging now a real source of new results. Degradation processes such as hydrogen embrittlement in 316L austenitic stainless steel call for accurate, multi-scale characterization of both the microstructure and the fracture surface. SEM analysis has traditionally relied on manual inspection by experts, which is slow, partly subjective, and a bottleneck for high-throughput materials development and quality control. Deep learning (DL), and CNNs in particular, has transformed automated computer vision in biomedicine and autonomous driving. Metallography has been slower to adopt these methods for three reasons: large, well-annotated open-access datasets are lacking, micrographs carry a high density of information, and the decisions of neural networks are hard to interpret (the ``black-box'' problem). If automated pipelines are to produce reproducible results that follow the FAIR (Findable, Accessible, Interoperable, and Reusable) data principles, the models have to be validated on material regions they have not seen before. We therefore build a data processing and machine learning pipeline designed specifically for region-held-out evaluation of SEM-based HE detection in 316L, and use it to examine how the models perform, how reliable they are, and how their decisions depend on architecture when they meet distinct, unseen microstructural domains~\cite{stiefel2024,durmaz2021,oakdenrayner2020}.

Deep learning architectures can match or exceed human experts in automated microstructure evaluation. Aggregate metrics reported on established benchmark datasets can still overstate how useful and safe a model is for engineering. One cause is data leakage, which can arise from hidden stratification of training and test sets, where small but important structural subsets or spatial correlations are buried inside coarse labels. In micrograph analysis, a model may rely on background artifacts, specimen-preparation textures or local imaging features rather than on the underlying damage mechanism. The large, dominant parts of a test set can then hide very poor performance on rare but safety-critical features such as early-stage micro-cracking or local phase transitions. For HE validation in 316L stainless steel, this kind of leakage gives a false sense of security that breaks down as soon as the model is applied to a new region of material. A region-held-out evaluation narrows the gap between optimistic laboratory benchmarks and reliable real-world failure detection. Physically separating training and evaluation data exposes hidden stratification, prevents leakage, and forces the network to learn from genuine metallurgical features rather than from region-specific ones.

Previous machine learning work has focused on corrosion and \Htwo{} detection. Localized corrosion, stress corrosion cracking (SCC) and hydrogen embrittlement (HE) are catastrophic failure modes of critical structural alloys in safety-critical applications such as nuclear containment, aerospace components and hydrogen storage infrastructure~\cite{varoquaux2022,aich2026}. Early-stage microstructural weaknesses, including micro-fissures, localized pitting and subsurface damage, have so far been identified only by labor-intensive and subjective manual inspection of SEM micrographs. Advances in artificial intelligence and computer vision (AI-CV) have made it possible to apply CNN-based models such as ResNet50, EfficientNetV2 and YOLO architectures to automated defect classification and semantic segmentation. These models have clear limitations in specialized metallurgical settings. Most are trained on high-contrast natural images or on specific medical imaging tasks, and when deployed on 316L stainless steel microstructures they give inaccurate boundary delineation and highly ambiguous labels. Much of the existing literature also uses datasets split statically and at random across the whole material domain. This ignores the highly localized and time-dependent spatial distribution of hydrogen-induced damage and produces overly optimistic metrics that do not carry over to unseen material regions. Because comprehensive metallurgical datasets are scarce and conventional validation is often inadequate, we built a dedicated microscopy dataset and propose a region-held-out evaluation paradigm. Decoupling localized anomaly segmentation from global defect severity assessment shows whether a model can detect and segment the heterogeneous signatures of hydrogen embrittlement and corrosion in fully independent material regions~\cite{aminudin2025,zhao2025}.

In this study, we combine SEM image analysis with a region-held-out, Leave-One-Region-Out (LORO) cross-validation protocol that prevents spatial data leakage between training and test sets. Classifiers, including a convolutional neural network (CNN), are used to separate as-received (AR) from hydrogen-charged (\Htwo) specimens, and their performance is checked with a formal group-level permutation test. Finally, a CNN interpretability analysis based on Gradient-weighted Class Activation Mapping (Grad-CAM) is used to examine the fracture-surface microstructure of 316L stainless steel.

\section{Methodology}

Image classification was implemented as a Python pipeline built on PyTorch, scikit-learn and scikit-image. Filenames followed the convention \texttt{material\_condition\_region\_imageid}, and parsing them produced a metadata table of 143 SEM micrographs covering several alloys (316L, X65, 100Cr6, 304) and two exposure conditions (as-received, AR; hydrogen-charged, \Htwo). The primary classification task used the 316L subset: 31 images from 14 distinct regions (8 AR, 6 \Htwo). Region identity served as the leakage-prevention unit for every data split.

For self-supervised feature learning, a convolutional autoencoder was trained for 30 epochs on all 143 unlabeled images with a mean squared error (MSE) reconstruction loss, reaching a final loss of 0.0162. All images were resized to $96 \times 96$ pixels and converted to single-channel tensors before encoding. The trained encoder was then frozen and used to produce global embeddings (SSL\_global) for the 316L classification task. Texture descriptors were of two kinds. Local Binary Patterns (LBP)~\cite{ojala2002} were computed in their uniform variant (LBPU) with radius $r = 3$ pixels and $P = 24$ neighboring points, giving a 26-dimensional histogram. Gray-Level Co-occurrence Matrix (GLCM)~\cite{haralick1973} features were computed at distances $d = [1, 3, 5]$ pixels and four orientation angles ($0$, $\pi/4$, $\pi/2$, $3\pi/4$), with contrast, dissimilarity, homogeneity, energy, correlation and angular second moment as the derived statistics (72 dimensions). All texture features were computed on images resized to $96 \times 96$ pixels. Each descriptor was used on its own (LBP+SVM, GLCM+SVM) and in combination (LBP+GLCM+SVM, LBP+GLCM+RF). The classifiers were a Support Vector Machine with a radial basis function kernel (SVM-RBF)~\cite{cortes1995} and a Random Forest, both with \texttt{class\_weight = 'balanced'} to account for the class imbalance (8 AR vs.\ 6 \Htwo{} regions). The SSL\_global embeddings (64-dimensional) were classified with an $\ell_2$-regularized logistic regression ($C = 0.1$, \texttt{class\_weight = 'balanced'}). As a deep-learning baseline, a compact convolutional neural network (SmallCNN) was trained directly on the raw $96 \times 96$ image tensors. In each LORO fold, the SmallCNN was trained from scratch for 10 epochs on the training regions, using random horizontal and vertical flips, rotations of up to $15^\circ$ and brightness/contrast jitter, together with class-weighted cross-entropy and weighted random sampling. Because the autoencoder was pretrained without labels on all 143 images, the held-out images were seen during pretraining but never with their labels.

Models were evaluated with 14-fold Leave-One-Region-Out (LORO) cross-validation. Each fold withheld every image from a single region (1--4 images per fold), and predictions were pooled across folds into one confusion matrix per model. Performance was measured by balanced accuracy, F1 score, precision and recall, with emphasis on the \Htwo{} (minority) class because missed \Htwo{} cases matter most for safety. Statistical significance was assessed with a group-level permutation test~\cite{ojala2010,good2005}, applied to the best-performing model (LBP+SVM) and, for comparison, to LBP+GLCM+SVM. The region-to-label assignment was permuted (500 draws from the $\binom{14}{6} = 3003$ possible assignments), and the full LORO pipeline was re-run for each permutation. This produced a null distribution of balanced accuracy against which the observed score was compared ($p < 0.05$). For interpretability, Gradient-weighted Class Activation Mapping (Grad-CAM) was applied to a CNN trained on the entire 316L image set (no held-out split) to locate the image regions that most strongly separate AR from \Htwo{} surface morphologies. All experiments used the same random seed (seed = 42) so that they can be repeated exactly~\cite{neftci2019,chen2020}.

\section{Results and Discussion}

\subsection{Overview of LORO classification performance}
\label{sec:overview}

We evaluated six feature-classifier combinations with Leave-One-Region-Out (LORO) cross-validation over the 14 spatial regions (8 AR, 6 \Htwo) to see how well as-received (AR) and hydrogen-charged (\Htwo) 316L specimens can be told apart. Table~\ref{tab:loro} lists the balanced accuracy (BA) of each model together with the F1 score, precision and recall for the \Htwo{} class. The texture-based LBP+SVM performed best, with a balanced accuracy of $0.79 \pm 0.36$. This is well above the convolutional baseline (CNN, BA $= 0.67 \pm 0.45$), the combined texture descriptors (LBP+GLCM+SVM, $0.62 \pm 0.43$; LBP+GLCM+RF, $0.59 \pm 0.45$), the global embeddings (SSL\_global, $0.56 \pm 0.43$) and GLCM alone ($0.52 \pm 0.46$, close to the chance level of 0.50). LBP+SVM also gave the best F1 score (0.75) and precision (0.82) for the \Htwo{} class, with a recall of 0.69.

Two points stand out in this ranking. First, the best model was a classical pipeline: a hand-crafted texture descriptor (Local Binary Patterns, uniform variant, $r = 3$, $P = 24$) fed to an SVM with an RBF kernel~\cite{cortes1995}, which has far fewer trainable parameters than the convolutional or self-supervised architectures. This agrees with earlier findings that, at small sample sizes (here 31 images from 14 independent regions), low-capacity classical descriptors generalize more reliably, while deep representation-learning methods tend to overfit region-specific artifacts instead of condition-relevant texture cues~\cite{brigato2021,barz2020}. Second, adding GLCM features to LBP (LBP+GLCM+SVM) did not raise the balanced accuracy. It lowered it, from 0.79 to 0.62, which suggests that the extra GLCM dimensions were noisy or redundant and that the SVM could not fully suppress them, even with class-balanced weighting~\cite{cabrera2024}.

Taken together, these results indicate that grain-boundary and surface-texture statistics carry the most condition-discriminative signal in this dataset, and that a simple, low-variance classifier extracts that signal better than deeper or higher-dimensional representations. Given the size of the gap (0.79 against 0.50--0.67 for all other models), LBP+SVM is the natural choice for the rest of the analysis (Sections~\ref{sec:confusion}--\ref{sec:gradcam}). We keep the full comparison in Table~\ref{tab:loro} to show that this choice was reached empirically and was not made \textit{a priori}.

\begin{table}[htbp]
\centering
\caption{LORO cross-validation results (316L AR vs.\ \Htwo, 14-fold).}
\label{tab:loro}
\begin{tabular}{lcccc}
\toprule
Model & Balanced accuracy & F1 (\Htwo) & Precision (\Htwo) & Recall (\Htwo) \\
\midrule
LBP+SVM       & $0.79 \pm 0.36$ & 0.75 & 0.82 & 0.69 \\
CNN           & $0.67 \pm 0.45$ & 0.67 & 0.55 & 0.85 \\
LBP+GLCM+SVM  & $0.62 \pm 0.43$ & 0.52 & 0.60 & 0.46 \\
LBP+GLCM+RF   & $0.59 \pm 0.45$ & 0.50 & 0.55 & 0.46 \\
SSL\_global   & $0.56 \pm 0.43$ & 0.48 & 0.50 & 0.46 \\
GLCM+SVM      & $0.52 \pm 0.46$ & 0.36 & 0.44 & 0.31 \\
\bottomrule
\end{tabular}

\vspace{0.5em}
\begin{minipage}{0.92\textwidth}
\footnotesize Balanced accuracy is reported as mean $\pm$ standard deviation across the 14 LORO folds. F1, precision and recall are reported for the \Htwo{} (minority) class on pooled out-of-fold predictions. Chance-level balanced accuracy $= 0.50$.
\end{minipage}
\end{table}

\subsection{Pooled confusion matrix and \texorpdfstring{\Htwo{}}{H2} recall}
\label{sec:confusion}

\begin{figure}[htbp]
\centering
\includegraphics[width=0.6\textwidth]{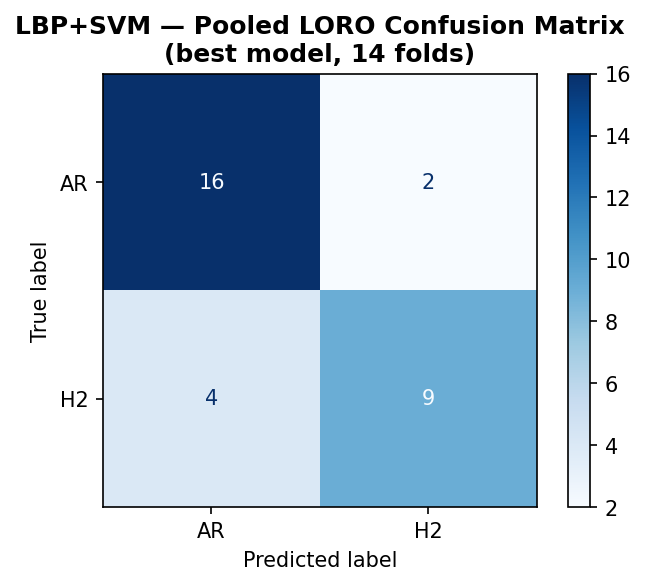}
\caption{Pooled confusion matrix for the LBP+SVM model under Leave-One-Region-Out cross-validation, aggregated across all 14 folds (31 images in total: 18 AR, 13 \Htwo). Diagonal cells are correct classifications; off-diagonal cells are misclassifications. The model reaches a specificity of 0.89 for AR and a recall of 0.69 for \Htwo, with 4 false negatives (\Htwo{} classified as AR) and 2 false positives (AR classified as \Htwo).}
\label{fig:confusion}
\end{figure}

To see where the errors behind the 0.79 balanced accuracy of LBP+SVM come from, we pooled the out-of-fold (OOF) predictions from all 14 LORO folds into a single confusion matrix covering all 31 images (18 AR, 13 \Htwo; Table~\ref{tab:confusion} and Figure~\ref{fig:confusion}).

\begin{table}[htbp]
\centering
\caption{Pooled confusion matrix (LBP+SVM, 31 images).}
\label{tab:confusion}
\begin{tabular}{lcc}
\toprule
 & Predicted AR & Predicted \Htwo \\
\midrule
Actual AR ($n = 18$)    & 16 & 2 \\
Actual \Htwo{} ($n = 13$) & 4  & 9 \\
\bottomrule
\end{tabular}
\end{table}

The model classified 16 of 18 AR images correctly (specificity $= 0.89$) and 9 of 13 \Htwo{} images (recall $= 0.69$). It missed 4 \Htwo{} images (false negatives) and flagged 2 AR images as \Htwo{} (false positives). The gap between specificity (0.89) and \Htwo{} recall (0.69) is the most important feature of this result in practice. For hydrogen-embrittlement screening, a false negative, meaning an \Htwo-exposed specimen that is declared AR even though it is degraded, is the more dangerous error. The four false negatives therefore set a practical limit on the model as a stand-alone screening tool: at present it misses roughly one in three \Htwo-exposed regions. At the same time, a recall of 0.69 obtained under a strict region-held-out protocol, where no image from the held-out region was ever seen in training, is a considerably more conservative and credible figure than an image-level split would give, since such a split can place images from the same region in both the training and test sets~\cite{hwang2025}. The fairly high precision (0.82) means that when the model predicts \Htwo, it is usually right. This points to a genuine descriptor of hydrogen exposure, a real ``texture signature,'' and not a spurious correlate of the small dataset. These results lead into the variance and significance analysis of LBP-based grain-boundary texture statistics in Sections~\ref{sec:variance}--\ref{sec:permutation}. That analysis shows that LBP is not yet a deployable diagnostic, but that an identifiable \Htwo{} signal is present under leakage-safe conditions.

\subsection{Fold-level variance and the cost of region-held-out evaluation}
\label{sec:variance}

\begin{figure}[htbp]
\centering
\includegraphics[width=\textwidth]{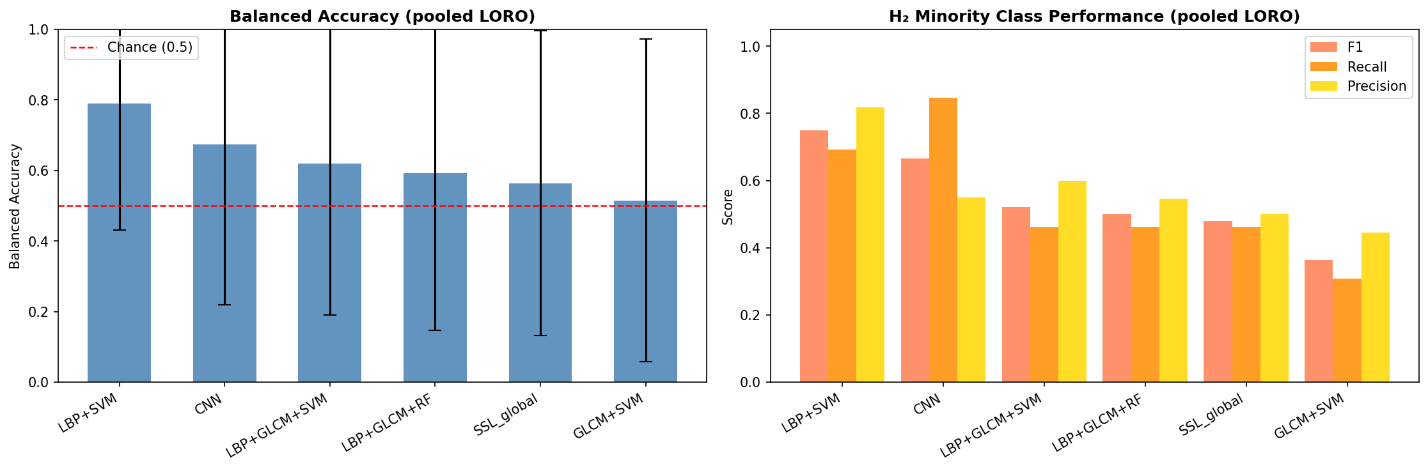}
\caption{Mean balanced accuracy ($\pm$ standard deviation) for each of the six model configurations under 14-fold Leave-One-Region-Out cross-validation. The chance level (BA $= 0.50$) is shown as a reference line. LBP+SVM has the highest mean balanced accuracy but also a large variance, a consequence of the small per-fold sample sizes that come with region-level evaluation.}
\label{fig:meanba}
\end{figure}

\begin{figure}[htbp]
\centering
\includegraphics[width=\textwidth]{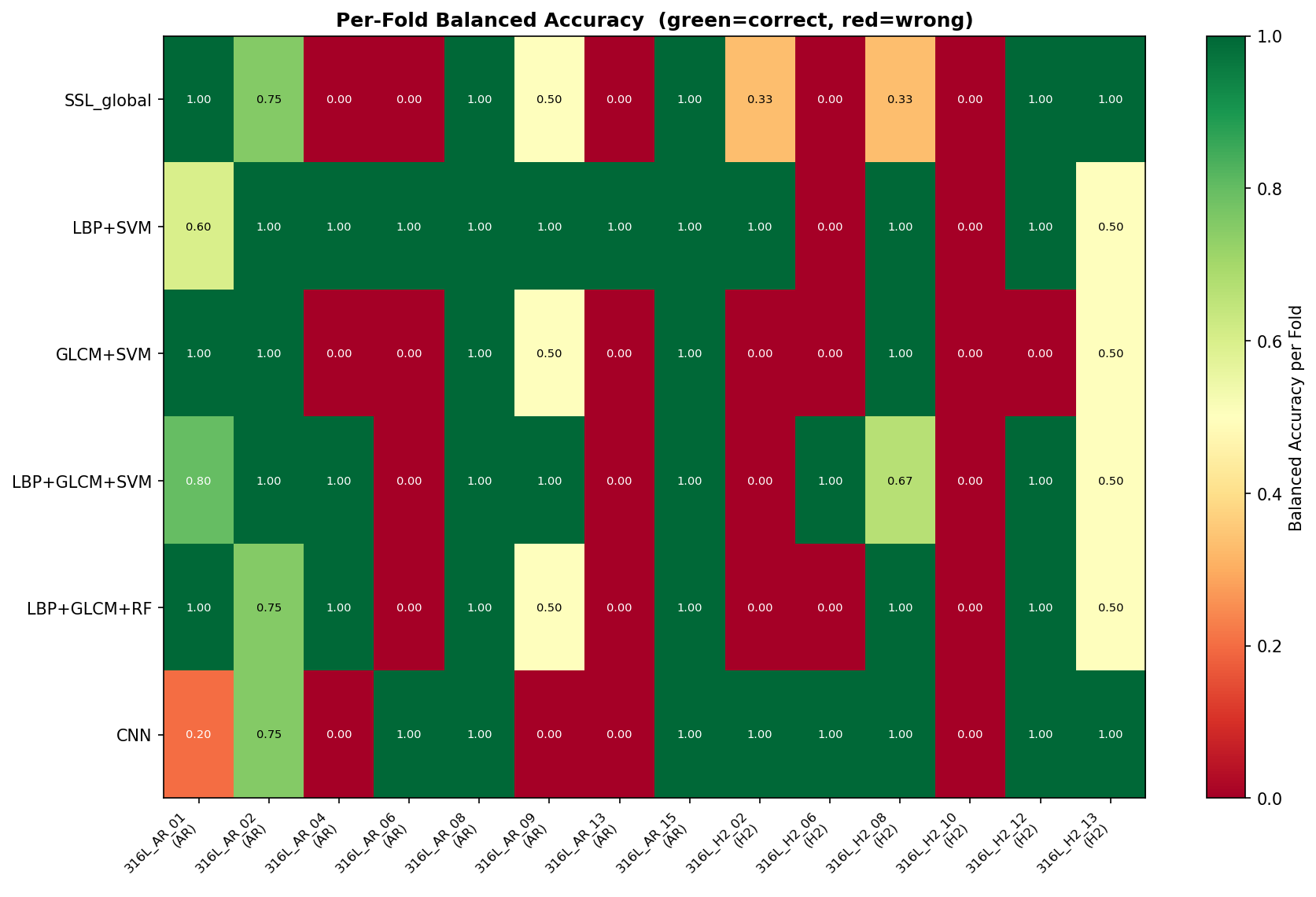}
\caption{Per-fold balanced accuracy of the six models across the 14 held-out regions. Darker and lighter cells indicate higher and lower balanced accuracy in that region. Regions with consistently low accuracy across several models are candidates for closer inspection.}
\label{fig:heatmap}
\end{figure}

The standard deviations of the aggregate results were large across folds, from $\pm 0.36$ for LBP+SVM to $\pm 0.46$ for GLCM+SVM. The aggregate figures therefore hide considerable variability across the 14 LORO folds. Figure~\ref{fig:meanba} shows the balanced accuracy of each model averaged over folds, against the chance baseline (BA $= 0.50$), and Figure~\ref{fig:heatmap} shows the balanced accuracy for each individual held-out region. This variance does not come from poor model design. It follows directly from the protocol, which spreads 31 images over 14 held-out regions. Each fold holds out only 1--4 images, so its balanced accuracy can take only a few discrete values (0, 0.5 or 1.0 for a two-image fold), and a single misclassified image can shift a fold's score by 0.5 or more. Figure~\ref{fig:heatmap} makes clear that even the best model (LBP+SVM) scores very well on most regions and poorly on only a few. This small group of ``hard'' regions is what inflates the overall standard deviation.

The finding also has a methodological implication that goes beyond this dataset. Studies that compare classifiers by splitting images, rather than regions, into training and test sets average over the same variance. They also let information leak between training and test data, for example when two images from the same region, with similar illumination, charging artifacts or local microstructural features, end up in different folds~\cite{basa2021}. Such evaluation designs usually report higher and more stable accuracies, but the reasons lie in shared region-level artifacts and not in a condition-discriminative signal. The variance reported here instead reflects the real difficulty of generalizing to an entirely new specimen region. It is a less flattering but more realistic estimate of how the model would perform in practice. We draw two lessons from this. First, the per-fold heatmap (Figure~\ref{fig:heatmap}) can identify regions that several models misclassify, and these may need to be re-examined as possible outliers or mislabeled cases. Second, a variance this large means that the significance of any single aggregate balanced-accuracy value cannot be judged from the mean alone. This is why we performed the permutation test described in Section~\ref{sec:permutation}.

\subsection{Statistical significance via permutation testing}
\label{sec:permutation}

\begin{figure}[htbp]
\centering
\includegraphics[width=\textwidth]{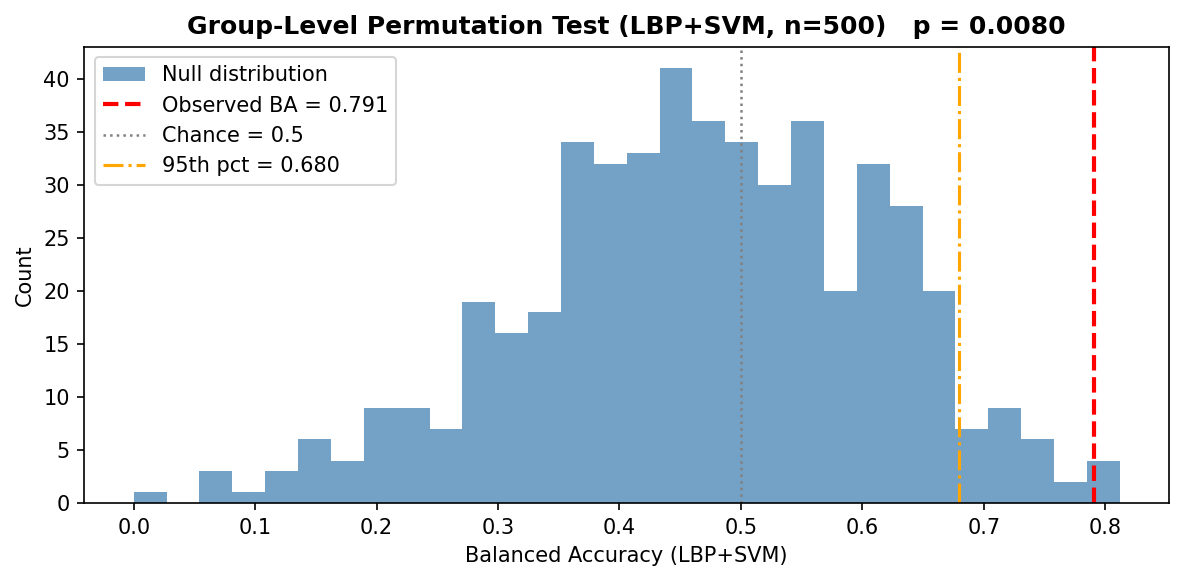}
\caption{Null distribution of balanced accuracy for the LBP+SVM model, generated from 500 permutations of the region-to-label assignment under the Leave-One-Region-Out protocol. The observed balanced accuracy (0.79, dashed red line) sits in the far right tail of the null distribution ($p = 0.008$). The chance level (0.50) and the 95th percentile of the null distribution are shown for reference.}
\label{fig:permutation}
\end{figure}

We applied a group-level permutation test to the best model from Section~\ref{sec:overview} (LBP+SVM, BA $= 0.79$) to check whether its LORO performance reflects a genuine condition-discriminative signal or is simply an effect of the small sample. The region-to-label assignment was permuted 500 times, drawing from the $\binom{14}{6} = 3003$ ways of choosing 6 \Htwo{} regions among 14, and the whole LORO pipeline was re-run for each permutation. The result is a null distribution of balanced accuracy (Figure~\ref{fig:permutation})~\cite{dheepak2024}.

The observed balanced accuracy of 0.7906 lay well above the null distribution: only 3 of the 500 permutations reached a balanced accuracy equal to or higher than the observed value, giving a permutation $p$-value of 0.008. This is significant at the conventional $p < 0.05$ level, so the performance of LBP+SVM under the region-held-out protocol is not the product of a chance arrangement of region labels. This answers the central methodological question of the study: do texture-based features carry a condition-discriminative signal that survives the removal of leakage? They do. Local Binary Pattern descriptors of grain-boundary and surface texture, combined with an SVM-RBF classifier, separate AR from \Htwo{} specimens at a level that differs significantly from random region-label assignment, with only 14 independent regions available.

The same permutation procedure applied to LBP+GLCM+SVM gives a useful contrast: its result was not significant (observed BA $= 0.62$, $p = 0.174$). The two runs share the evaluation method and the dataset and differ only in the added GLCM features. The GLCM features therefore contributed more noise than useful signal, in line with the feature-redundancy explanation proposed in Section~\ref{sec:overview}. The comparison also shows that the test is sensitive enough to pick up modest differences between feature representations (0.79 vs.\ 0.62 mean balanced accuracy) and can serve as a discriminating tool for model selection in small-sample SEM studies. Combined with the other results, the permutation test supports the main conclusion of this work: under a strict region-held-out approach, 316L SEM micrographs contain a real and statistically significant hydrogen-charging signature.

\subsection{Grad-CAM interpretability and physical plausibility}
\label{sec:gradcam}

\begin{figure}[htbp]
\centering
\includegraphics[width=\textwidth]{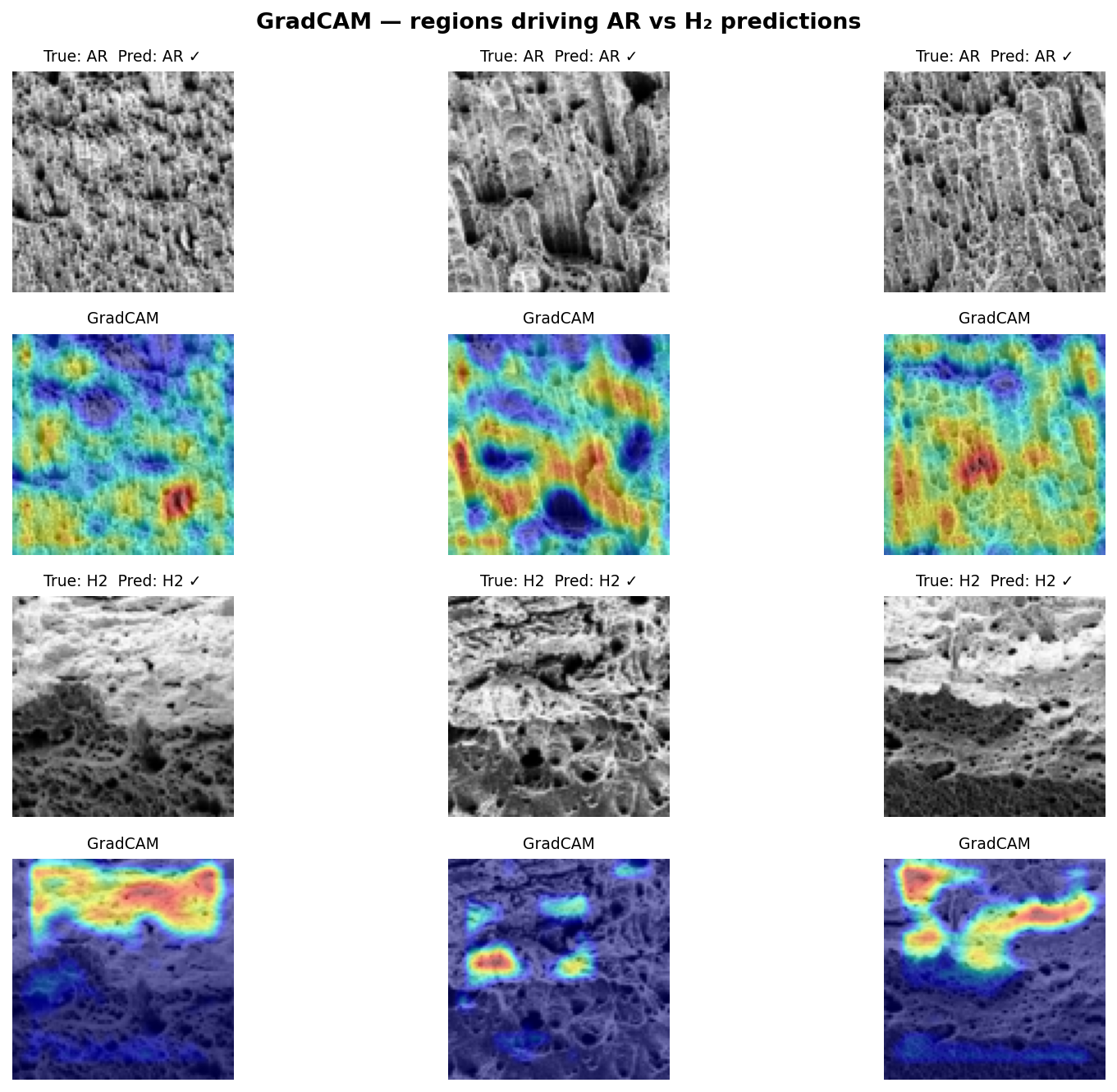}
\caption{Grad-CAM activation maps overlaid on representative SEM images of three AR and three \Htwo{} specimens, generated from a CNN trained on the full 316L dataset (20 epochs, no held-out split). Warmer colors mark regions with more influence on the model's prediction.}
\label{fig:gradcam}
\end{figure}

To obtain independent, non-statistical evidence on which visual features drive the classification, we trained a convolutional neural network (SmallCNN) on all 316L images (no held-out split, 20 epochs) and analyzed it with Gradient-weighted Class Activation Mapping (Grad-CAM)~\cite{qassim2018}. Activation maps were generated for a representative sample of three AR and three \Htwo{} images (Figure~\ref{fig:gradcam}). No single activation pattern was consistent across the sampled images. The activations did, however, tend to be localized within each micrograph rather than spread across the whole field of view. This suggests that the network responds to specific image features, such as grain-boundary networks and surface relief, and not to global properties like overall brightness or contrast, which would produce activation across the entire image. In the \Htwo{} images, the activation hot spots often coincided with areas whose surface texture looked different from that of the AR images. These areas corresponded qualitatively to literature descriptions of hydrogen-induced changes in grain-boundary cohesion and surface texture in austenitic stainless steels~\cite{cavaliere2025}.

\begin{figure}[htbp]
\centering
\includegraphics[width=\textwidth]{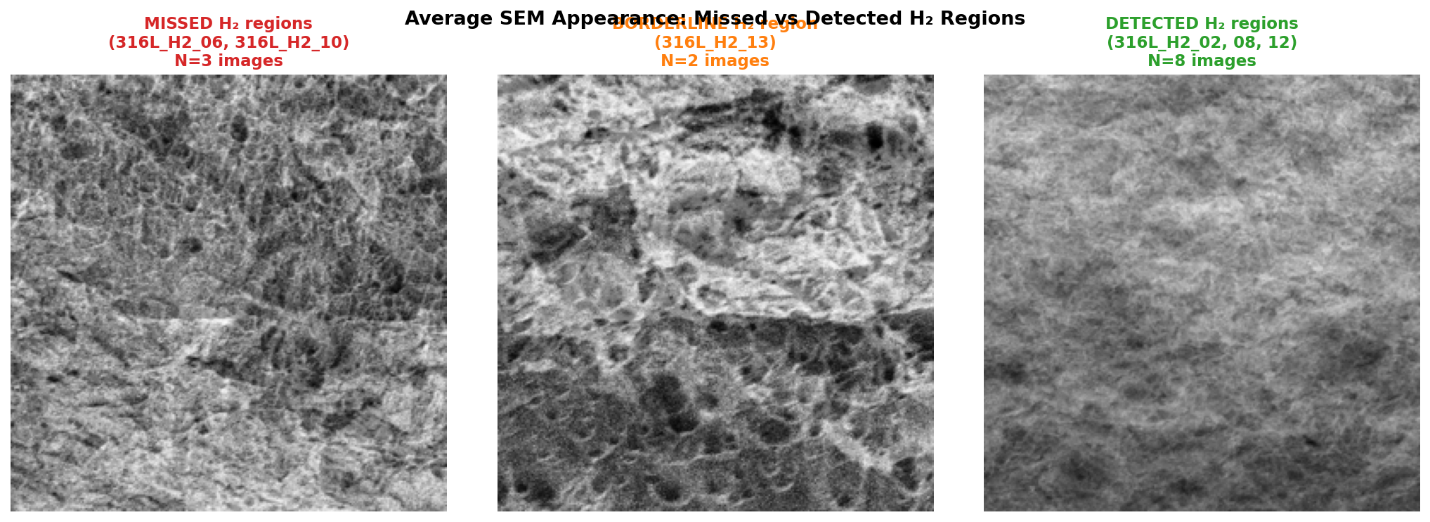}
\caption{Pixel-wise mean image computed across all \Htwo-labeled 316L specimens, showing consistent large-scale intensity and texture patterns associated with the hydrogen-charged condition.}
\label{fig:meanimg}
\end{figure}

\begin{figure}[htbp]
\centering
\includegraphics[width=\textwidth]{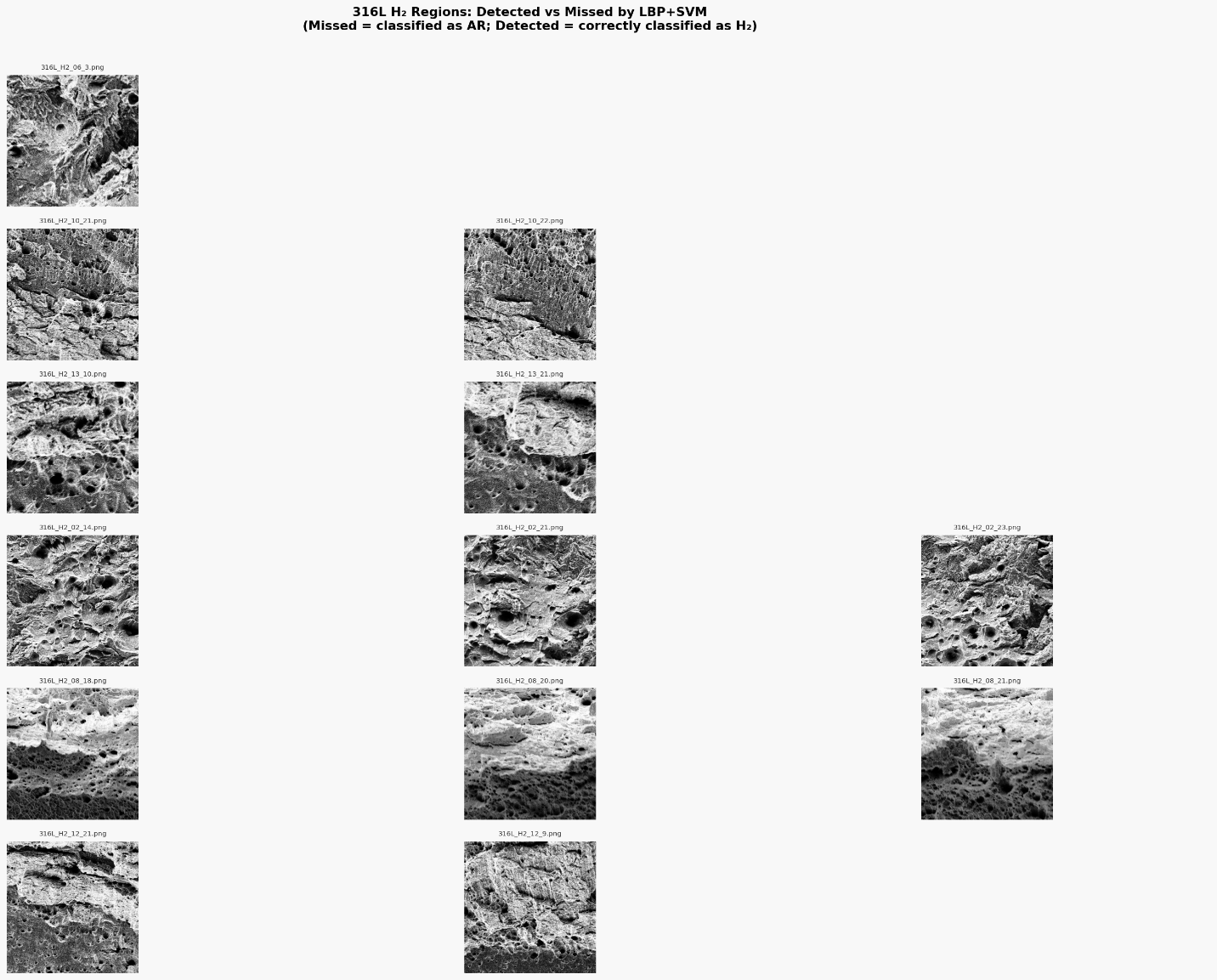}
\caption{Region-level comparison of texture characteristics between the AR and \Htwo{} specimen groups.}
\label{fig:regiontex}
\end{figure}

\begin{figure}[htbp]
\centering
\includegraphics[width=\textwidth]{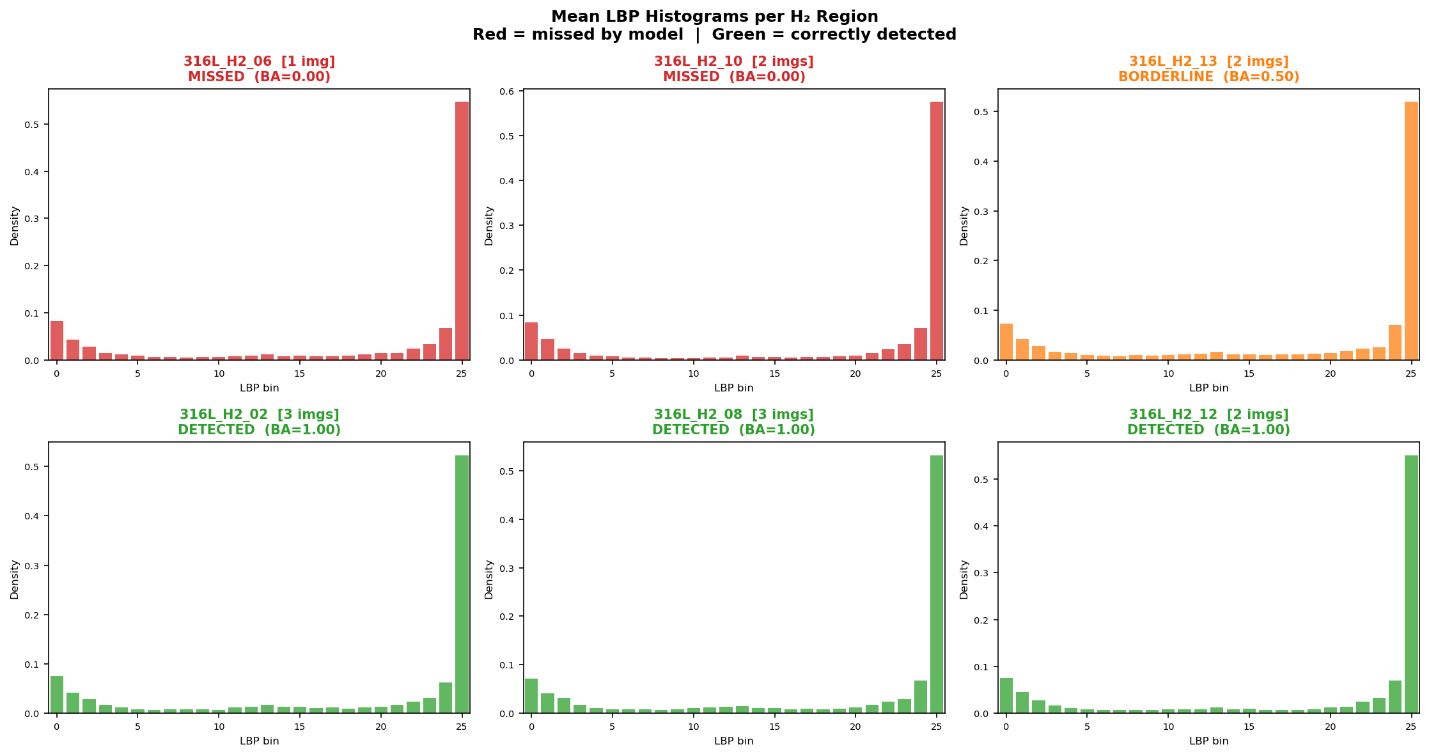}
\caption{Distribution of Local Binary Pattern histogram values for \Htwo-labeled images, showing the texture feature space used by the LBP+SVM classifier.}
\label{fig:lbphist}
\end{figure}

As a model-independent complement to Grad-CAM, we also show the pixel-wise mean intensity image across all \Htwo{} images (Figure~\ref{fig:meanimg}) and compare region-level texture statistics between the AR and \Htwo{} groups (Figure~\ref{fig:regiontex}). Because the LBP+SVM classifier works in the texture descriptor space, the distribution of LBP histogram values for the \Htwo{} images (Figure~\ref{fig:lbphist}) gives a direct view of what it uses. Together, these qualitative analyses serve two purposes. First, the spatial agreement between the Grad-CAM activations and known \Htwo{} damage morphology suggests that the classifier relies on features with a plausible physical basis rather than on confounds such as imaging artifacts or background. Second, the LBP histogram comparison (Figure~\ref{fig:lbphist}) links the best classical model (LBP+SVM, Section~\ref{sec:overview}) directly and interpretably to the image data. Because that model rests on an inspectable image representation, it answers a common criticism of ``black-box'' SEM classifiers. We stress that this interpretability analysis is qualitative and based on only six images. It is meant to support, not replace, the quantitative LORO and permutation-test results of Sections~\ref{sec:overview}--\ref{sec:permutation}, and it should be read as converging evidence rather than as evidence in its own right.

\section{Conclusion}

We evaluated a leakage-safe, region-held-out protocol for classifying hydrogen-charging effects (AR vs.\ \Htwo) in SEM micrographs of 316L stainless steel. Across six feature-classifier combinations tested with 14-fold Leave-One-Region-Out cross-validation, the best performance came from a classical texture descriptor: Local Binary Patterns paired with an SVM-RBF (LBP+SVM), with a balanced accuracy of 0.79, \Htwo{} recall of 0.69 and \Htwo{} precision of 0.82. It outperformed a convolutional neural network, self-supervised global embeddings and GLCM-based descriptors.

A group-level permutation test (500 permutations over the 3003 possible region-to-label arrangements) showed that the LBP+SVM performance is statistically significant ($p = 0.008$) and not due to a chance arrangement of the 14 specimen regions. Significance held despite the large fold-level variance that comes with a 31-image, 14-region dataset, because the permutation test accounts for that variance directly instead of relying on aggregate accuracy alone. Grad-CAM analysis of a CNN trained on the full 316L dataset gave independent, qualitative support: the activations concentrated on localized surface and grain-boundary features, consistent with literature descriptions of hydrogen-induced morphological changes.

Taken together, the results suggest that:
\begin{enumerate}[label=(\alph*)]
    \item Simple, low-parameter texture descriptors can outperform deep and self-supervised representations in small-sample SEM classification tasks, in line with earlier findings in the wider small-data machine learning literature.
    \item Region-held-out evaluation combined with group-level permutation testing gives a statistically rigorous and leakage-safe framework for validating SEM-based condition classifiers. The approach is not tied to this dataset and could help address known reproducibility concerns in SEM-based machine learning studies.
    \item Even with only 31 images, a significant and mechanistically plausible hydrogen-charging signal exists at the texture level, which makes this approach a workable starting point for larger studies.
\end{enumerate}

Several extensions follow naturally. Adding regions and specimens would reduce the fold-level variance and increase confidence in the reported metrics. The protocol could be extended to the other alloy systems (X65, 100Cr6, 304), for which pilot data exist but are not yet large enough for region-held-out evaluation. It would also be worth testing whether combining LBP features with Grad-CAM-guided regions of interest improves \Htwo{} recall, the most safety-critical metric for any future screening application.

\section*{Author Contribution Statement}
MA conceptualized and developed the methodological framework, including algorithm design and implementation, experimental design, model training and evaluation, and performance analysis, and prepared the original draft of the manuscript. AS supervised the research and contributed to reviewing and editing the manuscript. NAN provided computational resources and contributed to reviewing and editing the manuscript. MY and HZ prepared and edited the figures and tables. MZS contributed to reviewing, editing, proofreading and language correction. All authors reviewed and approved the final version of the manuscript.

\section*{Data Availability Statement}
The SEM micrographs analyzed in this study are available from the corresponding
author on reasonable request. The metadata convention used to define specimen
regions (\texttt{material\_condition\_region\_imageid}), which determines the
grouping for Leave-One-Region-Out cross-validation, is documented in the code
repository.

\section*{Code Availability Statement}
The code used for feature extraction, self-supervised pretraining,
Leave-One-Region-Out cross-validation, group-level permutation testing and
Grad-CAM analysis is openly available at
\url{https://github.com/cmsawais/leakage-safe-hydrogen-embrittlement-ml}
under the MIT license~\cite{awais2026code}. The version used to produce the
results reported here is tagged as release \texttt{v1.0}.
All experiments use a fixed random seed (seed = 42) and were run on CPU with
Python~3.13, PyTorch~2.7.0 and scikit-learn~1.6.

\section*{Acknowledgments}
No financial support was received for this project.

\section*{Declaration of Competing Interest}
The authors declare that they have no known competing financial interests or personal relationships that could have influenced the work reported in this paper.

\bibliographystyle{unsrtnat}
\bibliography{references}

\end{document}